\documentclass[sigconf,natbib=true]{acmart}

\usepackage{bm}
\usepackage{multirow}
\usepackage{caption,subcaption}
\usepackage{tabularx}
\usepackage{float}
\usepackage{verbatim}

\usepackage{amsmath}

\usepackage[capitalize]{cleveref}
\crefname{section}{Sec.}{Secs.}
\Crefname{section}{Section}{Sections}
\Crefname{table}{Table}{Tables}
\crefname{table}{Tab.}{Tabs.}
\AtBeginDocument{%
  \providecommand\BibTeX{{%
    \normalfont B\kern-0.5em{\scshape i\kern-0.25em b}\kern-0.8em\TeX}}}

\setcopyright{acmlicensed}
\copyrightyear{2018}
\acmYear{2018}
\acmDOI{XXXXXXX.XXXXXXX}

\acmConference[Conference'17]{}{July 2017}{Washington D.C., USA}
\acmISBN{978-1-4503-XXXX-X/18/06}

\begin{document}

\title{Event-aware Video Corpus Moment Retrieval}


\author{Danyang Hou}
\affiliation{%
  \institution{CAS Key Laboratory of AI Security, Institute of Computing Technology, Chinese Academy of Sciences}
  \institution{University of Chinese Academy of Sciences}
  \city{Beijing}
  \country{China}}
\email{houdanyang18b@ict.ac.cn}

\author{Liang Pang}
\authornote{Corresponding author}
\affiliation{%
  \institution{CAS Key Laboratory of AI Security, Institute of Computing Technology, Chinese Academy of Sciences}
   \institution{University of Chinese Academy of Sciences}
  \city{Beijing}
  \country{China}}
\email{pangliang@ict.ac.cn}

\author{Huawei Shen}
\affiliation{%
  \institution{CAS Key Laboratory of AI Security, Institute of Computing Technology, Chinese Academy of Sciences}
  \institution{University of Chinese Academy of Sciences}
  \city{Beijing}
  \country{China}}
\email{shenhuawei@ict.ac.cn}

\author{Xueqi Cheng}
\affiliation{%
  \institution{CAS Key Lab of Network Data Science and Technology, Institute of Computing Technology, Chinese Academy of Sciences}
  \institution{University of Chinese Academy of Sciences}
  \city{Beijing}
  \country{China}}
\email{cxq@ict.ac.cn}



\begin{abstract}

Video Corpus Moment Retrieval (VCMR) is a practical video retrieval task focused on identifying a specific moment within a vast corpus of untrimmed videos using the natural language query.
Existing methods for VCMR typically rely on frame-aware video retrieval, calculating similarities between the query and video frames to rank videos based on maximum frame similarity. 
However, this approach overlooks the semantic structure embedded within the information between frames, namely, the event, a crucial element for human comprehension of videos.
Motivated by this, we propose EventFormer, a model that explicitly utilizes events within videos as fundamental units for video retrieval. The model extracts event representations through event reasoning and hierarchical event encoding. The event reasoning module groups consecutive and visually similar frame representations into events, while the hierarchical event encoding encodes information at both the frame and event levels. We also introduce anchor multi-head self-attenion to encourage Transformer to capture the relevance of adjacent content in the video.
The training of EventFormer is conducted by two-branch contrastive learning and dual optimization for two sub-tasks of VCMR. Extensive experiments on TVR, ANetCaps, and DiDeMo benchmarks show the effectiveness and efficiency of EventFormer in VCMR, achieving new state-of-the-art results.
Additionally, the effectiveness of EventFormer is also validated on partially relevant video retrieval task.

\end{abstract}

\ccsdesc[500]{Information systems~Video search}

\keywords{Video Corpus Moment Retrieval, Video Retrieval, Event Retrieval}


\maketitle

\begin{figure}[!htbp]
    \begin{center}
    \includegraphics[width=0.48\textwidth]{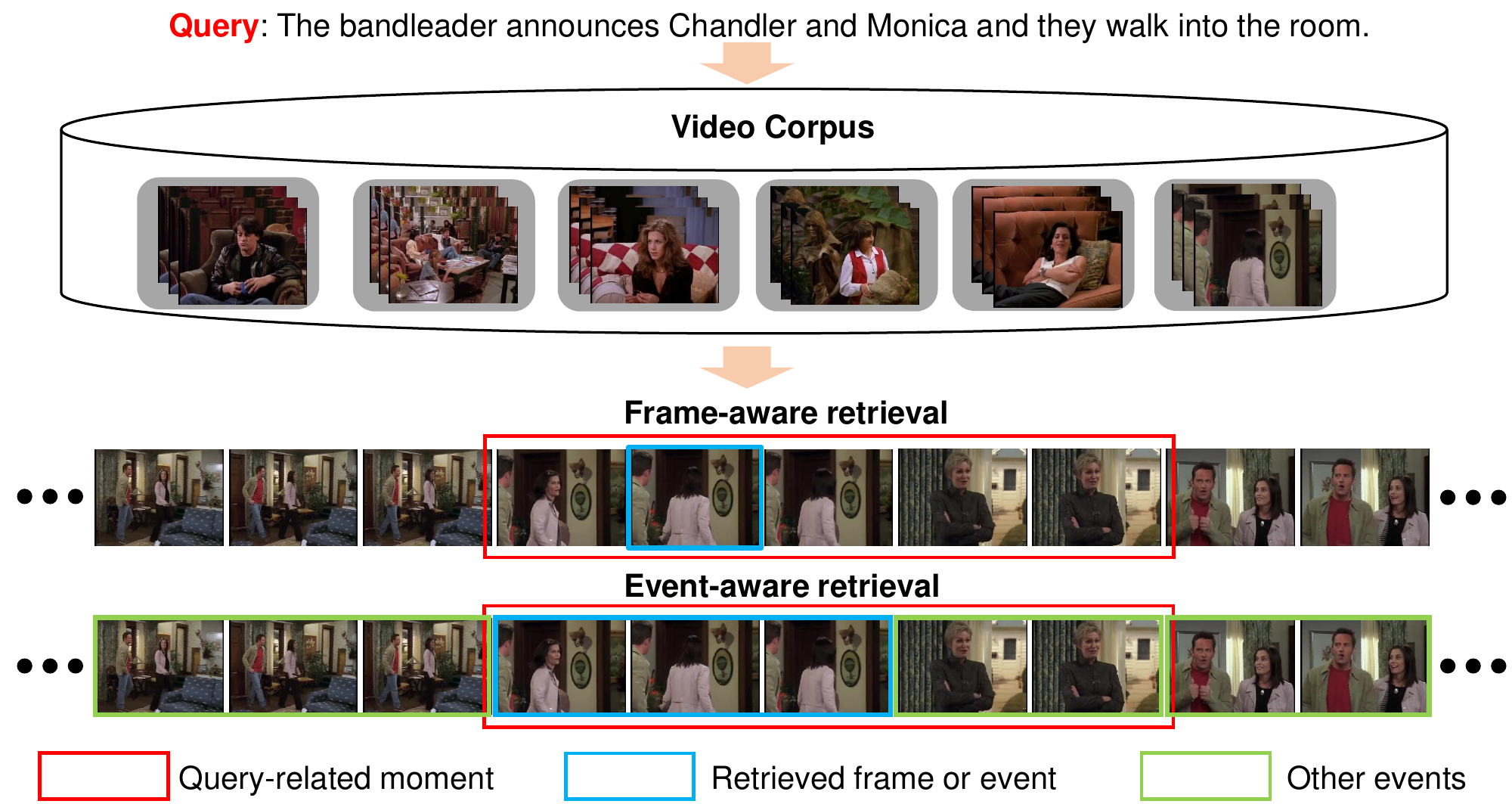}
    \end{center}
    \caption{In VCMR, the relevant part corresponding to the query is the moment. While the frame-aware method utilizes frames for retrieval, our event-aware approach adopts events as the retrieval unit, ensuring a more comprehensive capture of moment information.}
    \label{fig:teaser}
\end{figure}

\begin{table}[]
\caption{The overlap between the moments predicted by models or the extracted events with ground truth moments. The metric is the ratio of predicted moments to ground truth moments with an IoU greater than 0.5 or 0.7.}
\begin{tabular}{lll}
\hline
Model            & IoU=0.5       & IoU=0.7 \\
\hline
XML~\cite{lei2020tvr} (ECCV'20)              & 29.56         & 13.05        \\
ReLoCLNet~\cite{zhang2021video} (SIGIR'21)    & 31.65         & 14.80        \\
HERO~\cite{li2020hero} (EMNLP'20)            & 32.2          & \textbf{15.30}        \\
\hline
Event & \textbf{35.04}         & 14.21        \\
\hline
\end{tabular}
\label{tab:event_validation}
\end{table}

\section{Introduction}

With the widespread use of mobile devices and video-sharing applications, online video content has surged to unprecedented levels, encompassing extensive, untrimmed content, including TV series and instructional videos.
An advanced video retrieval system should efficiently pinpoint specific moments within a vast corpus for users, be it a classic shot from a movie or a crucial step in an instructional video, thereby minimizing the user's browsing time.
Addressing this need, the recently proposed Video Corpus Moment Retrieval task (VCMR)~\cite{temporal,lei2020tvr} requires retrieving semantically relevant video moments from a corpus of untrimmed videos by a natural language query, where the moment is a continuous temporal segment.

A distinctive feature of VCMR, setting it apart from typical text-to-video retrieval, lies in the nature of video relevance to the query. Unlike trimmed videos in text-to-video retrieval~\cite{chen2011collecting}, where the entire video aligns with the text query, the untrimmed video involves only a small part~(relevant moment)~of the content being related to the query, shown in~\Cref{fig:teaser}. The existing works~\cite{lei2020tvr,li2020hero,zhang2021video} employ frame-aware video retrieval to capture the partial relevance between the query and video. This entails calculating the similarity between the query and all video frames in the corpus, ranking the videos based on the maximum similarity of frames within each video. However, these works overlook the semantic structure embedded within the information between video frames, i.e. event. Cognitive science research~\cite{tversky2013event} suggests that human perception of visual information primarily revolves around the concept of events, with event information being the most fundamental unit of visual perception for humans. In the realm of videos, a sequence depicting a consistent action, object, or environment is termed an event~\cite{shou2021generic}, comprised of frames that are both similar and consecutive. Employing frames as a unit for video retrieval contains less information compared to human cognitive habits. While the event may not overlap with relevant moment exactly, it covers more complete information than a frame, shown in~\Cref{fig:teaser}. 

To further evaluate the helpfulness of the event for the VCMR task, we measure the overlap between the events extracted using the unsupervised method in~\cite{kang2022uboco} from the video and the ground truth moment. The results on the TVR validation set are shown in~\Cref{tab:event_validation}. The model results are based on the predicted moments, and the event extraction results are derived from video events with the highest overlap with the correct moment~(the ideal case). Notably, with the threshold set at 0.5, the optimal extracted events outperform the predicted moments of all models. Given that the events are extracted without any training, these results highlight the utility of event information in video for VCMR. If the event can be utilized effectively, it will enhance the accuracy of retrieval. Retrieval efficiency will also increase because fewer units reduce the amount of computation in retrieval.

However, the frame-aware method, which simply encodes the frame representations by Transformer, struggles to utilize event for retrieval. There are three main reasons. (1) \textbf{The event information is not explicitly extracted from the frames}. Although contextual relevance is captured using Transformer, each frame expresses more information from itself, posing challenges in capturing the overall information of an event.  Hence, directly using frame as event is insufficient. (2) \textbf{It lacks event-level information interaction}. Events encapsulate more comprehensive semantic information, and strong semantic associations typically exist between events, as seen in examples such as two correlative steps in an instructional video. (3) \textbf{The attention of model is not adequately concentrated}. The range of attention in vanilla Tansformer~\cite{vaswani2017attention} is the entire video. But not all content in the untrimmed informative video is relevant. The most relevant content tends to be intra-event, i.e., adjacent. 

To this end, we propose EventFormer to explicitly leverage event information to help VCMR. The model contains two main components for event learning: event reasoning and hierarchical event encoding. The event reasoning module plays a pivotal role in extracting event information from the video based on the frame representation. In reference to the works on generic event boundary detection~\cite{shou2021generic}, we introduce three event extraction strategies, contrastive convolution, Kmeans, and window, aimed at aggregating visually similar and consecutive frames as event. The hierarchical event encoding module captures interactions not only at the frame level but also at the event level, obtaining a more semantically relevant representation of events. To encourage the model to focus attention on adjacent content, anchor multi-head self-attention is introduced to augment Transformer. 
VCMR task includes two sub-tasks: video retrieval~(VR) and single video moment retrieval~(SVMR). In VR, the objective is to retrieve the most pertinent untrimmed video from a large corpus using a natural language description as query, while SVMR focuses on pinpointing the start and end times of the relevant moment within the retrieved video.
For the two subtasks, EventFormer adopts distinct training strategies. Specifically, two-branch contrastive learning and dual optimization. Both strategies essentially integrate frame and event into the training process.

We evaluate our proposed EventFormer on three benchmarks, TVR~\cite{lei2020tvr}, ANetCaps~\cite{caba2015activitynet}, and DiDeMo~\cite{anne2017localizing}. The results show the effectiveness and efficiency of EventFormer, achieving new state-of-the-art results.
Additionally, we validate the effectiveness of our model in the partially relevant video retrieval (PRVR) task.

Our main contributions are as follows:
\begin{itemize}
\item We propose an event-aware model EventFormer for VCMR, motivated by human perception for visual information.
\item We adopt event reasoning and hierarchical event encoding for event learning, and anchor multi-head self-attention to enhance close-range dependencies. 
\item Experiments on three benchmarks show the effectiveness and efficiency, achieving new state-of-the-art results on VCMR. We also validate the effectiveness of the model in PRVR task.
\end{itemize}

\section{Related work}
In this section, we first introduce works on two related tasks, text-to-video retrieval and natural language video localization. Then we review works on VCMR. Finally, we present works on generic event boundary detection. 

\noindent \textbf{Text-to-video retrieval} Similar to VR, text-to-video retrieval aims to find relevant videos from a corpus based on a natural language query. However, the distinction lies in the nature of query-video relevance. In text-to-video retrieval, the video is trimmed to precisely match the entire content of the video with the text query. 
Text-to-video retrieval methods are broadly categorized into two types: two-tower models~\cite{bain2021frozen, gabeur2020multi, ge2022bridging, ging2020coot, liu2019use, miech2020end, rouditchenko2020avlnet, xu2021videoclip} and one-tower models~\cite{fu2021violet, lei2021less, sun2019videobert, xu2021vlm, chen2020fine, han2021fine, wang2021t2vlad}. Two-tower models utilize separate encoders for obtaining video and query representations, employing a simple similarity function like cosine to measure relevance. These methods are efficient due to decomposable computations of query and video representations. On the other hand, one-tower models leverage cross-modal attention~\cite{Bahdanau_attention, vaswani2017attention} for deep interactions between query and video, enhancing retrieval accuracy. Some works~\cite{miech2021thinking, liu2021inflate, yu2022cross, lei2022loopitr} combine the strengths of both methods by employing a two-tower model for fast retrieval of potentially relevant videos in the initial stage, followed by a one-tower model to accurately rank the retrieved videos in the subsequent stage.

\noindent \textbf{Natural language video localization} The objective of the natural language video localization task is to pinpoint a moment semantically linked to the query. This task bears similarities to SVMR and can be viewed as a specialized case of VCMR, wherein the corpus comprises only one video, and the video must contain the target moment. Early works  can be broadly classified into two categories: proposal-based~\cite{liu2018cross,xu2019multilevel,chen2019semantic,xiao-etal-2021-natural,chen2018temporally, zhang2019man,zhang2021multi, liu2021context} models and proposal-free~\cite{yuan2019find,chen2020rethinking,zeng2020dense,li2021proposal,ghosh2019excl,chen2019localizing,zhang2020span} models.
In proposal-based methods, initial steps involve generating moment proposals as candidates, followed by ranking these proposals based on the similarity between the query and the proposals. On the other hand, proposal-free methods take a direct approach by predicting the start and end positions of the target moment in the video based on the query. 
Drawing inspiration from the success of Transformer, particularly in object detection tasks such as DETR~\cite{carion2020end} (DEtection TransfomeR), recent works propose DETR-based methods~\cite{lei2021detecting, moon2023query, liu2022umt, cao2021pursuit} for moment localization. These approaches simplify the post-processing of previous predictions into an end-to-end process.

\noindent \textbf{Video corpus moment retrieval} VCMR is first proposed by  Escorcia et al.~\cite{temporal}, introducing VR task on top of natural language video localization, with benchmarks derived from localization datasets such as ANetCaps. Zhang et al.~\cite{lei2020tvr} propose a dataset for VCMR, where the videos provide subtitles. Similar to the taxonomy applied to text-to-video retrieval, existing VCMR works fall into one-tower, two-tower, and two-stage methods. One-tower~\cite{zhang2020hierarchical, yoon2022selective} and two-tower methods~\cite{lei2020tvr, zhang2021video, li2020hero}, essentially treated as one-stage approaches, address VCMR as a multi-task problem, utilizing a shared backbone model with distinct heads for VR and SVMR. HAMMER~\cite{zhang2020hierarchical} is the first one-tower model with hierarchical fine-grained cross-modal interactions. SQuiDNet~\cite{yoon2022selective} utilizes causal inference to avoid the model learning bad retrieval biases. The two-tower method demonstrates superior retrieval efficiency, especially when dealing with numerous videos in the corpus. To capture partial relevance in VR, frame-aware retrieval methods are commonly employed. XML~\cite{lei2020tvr} is a pioneering work in VCMR using frame-aware retrieval, followed by enhancements in ReLoCLNet~\cite{zhang2021video}, leveraging contrastive learning. Li et al.~\cite{li2020hero} introduces HERO, a video-language pre-trained model, significantly improving overall performance. The two-stage method combines one-tower and two-tower approaches, utilizing the two-tower model for VR to quickly retrieve video and the one-tower model for SVMR to precisely localize moment. CONQUER~\cite{hou2021conquer}, DMFAT~\cite{zhang2023video} and CKCN~\cite{chen2023cross} are two-stage models that employ HERO for video retrieval and propose one-tower models as moment localizer. CONQUER introduce a moment localizer based on context-query attention (CQA)\cite{yu2018qanet}. DMFAT innovates with multi-scale deformable attention for multi-granularity feature fusion. And CKCN introduces a calibration network to refine important modality features.  Our model also adopts a two-stage approach, differing by integrating an event-aware retrieval strategy. 
Recently, Dong et al.~\cite{dong2022partially} introduces a new task partially relevant video retrieval~(PRVR) which is a weakly supervised version of VR, where the relevant moment is not provided.

\noindent \textbf{Generic event boundary detection}  Generic event boundary detection (GEBD)~\cite{shou2021generic} is a video understanding task designed to identify boundaries, dividing the video into several meaningful units that humans perceive as events. Typically, the frames within an event exhibit visual similarity and continuity, with event boundaries aligning with changes in action, subject, and environment. The task provides supervised and unsupervised settings, where the unsupervised setting is suitable to be generalized across various video understanding scenarios. UBoCo~\cite{kang2022uboco} is a representative work for unsupervised GEBD that leverages contrastive convolution to identify frames with drastic visual variations as event boundaries from the temporal self-similarity matrix (TSM) of video frames. We integrate the method into the event reasoning of the proposed EventFormer and implement two other strategies.

\begin{figure*}
    \begin{center}
    \includegraphics[width=0.9\textwidth]{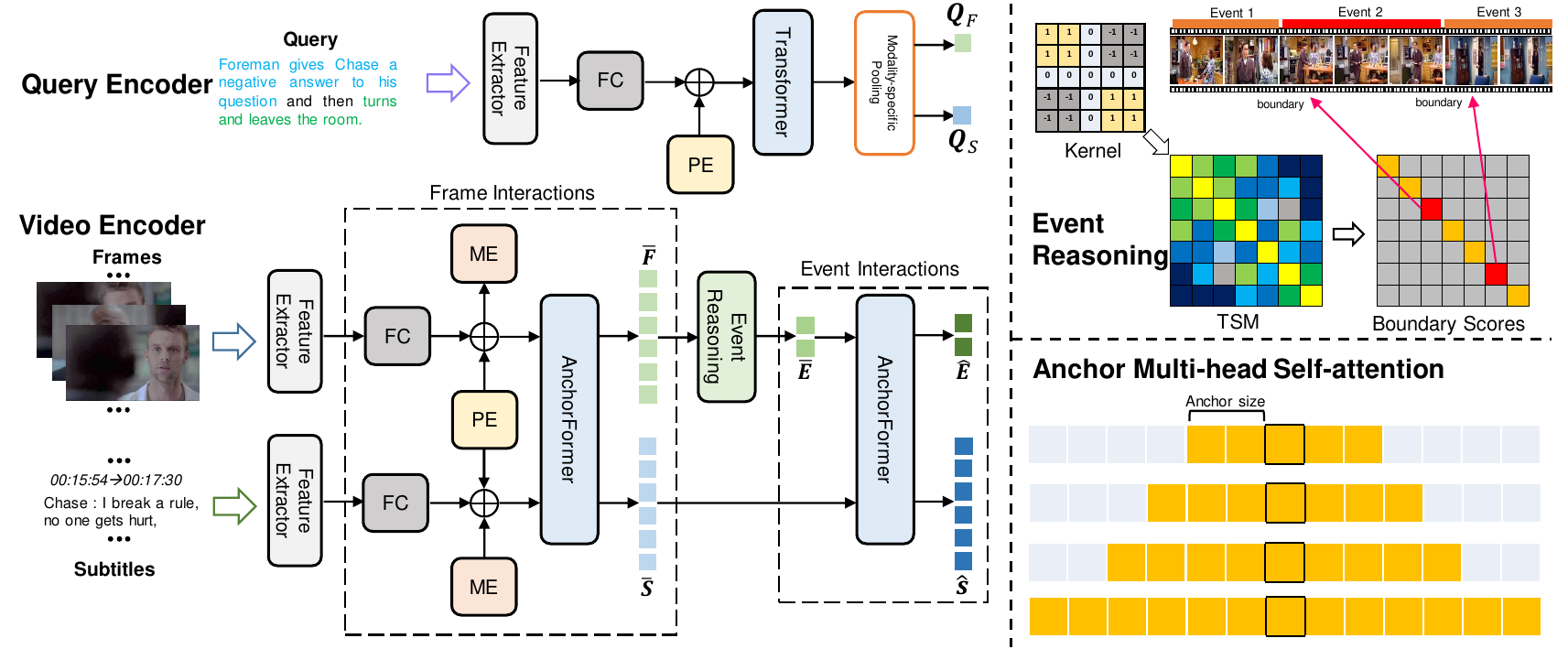}
    \end{center}
    \caption{Video retriever: the hierarchical encoding of events involves interactions at both the frame and event levels, where the events are extracted by event reasoning module and Transformer for frames or events is augmented with anchor attention. }
    \label{fig:retriever}
\end{figure*}

\section{Method}
In this section, we detail the proposed event-aware retrieval model EventFormer for VCMR task. We first formulate VCMR task and the sub-tasks in~\Cref{sec:task_formulation}. Then, we describe the feature extraction of video and query in~\Cref{sec:feature_extraction}. Next, we introduce two main modules of EventFormer video retriever and moment localizer in~\Cref{sec:video_retriever} and~\Cref{sec:moment_localizer} respectively. Finally, we present training and inference of model on VCMR task in~\Cref{sec:train_inference}.

\subsection{Task Formulation}
\label{sec:task_formulation}
Given a video corpus $\mathcal{V} = \{v_1, v_2,...,v_M\}$, the goal of VCMR is to retrieve the most relevant moment $m_*$ using a natural language query $q=\{w^1,w^2,...,w^L\}$ which consists of a sequence of words. The retrieval can be formulated as :
\begin{equation}
\label{equ:goal}
m_* = \mathop{\rm argmax}\limits_{m} P(m|q, \mathcal{V}).
\end{equation}


VCMR can be decomposed into two sub-tasks, VR and SVMR.
The goal of VR is to find the video  $v_*$ that potentially contains the target moment from the corpus:
\begin{equation}
\label{equ:goal_vr}
v_* = \mathop{\rm argmax}\limits_{v} P(v|q).
\end{equation}
And SVMR aims to localize moment from the retrieved video:
\begin{equation}
\label{equ:goal_svmr}
m_* = \mathop{\rm argmax}\limits_{m} P(m|v_*,q),
\end{equation}
where the predicted moment is decided by the start and end times:
\begin{equation}
\label{equ:st_ed_prob}
P(m|v_*,q) = P(\tau_{st}|v_*,q) \cdot P(\tau_{ ed}|v_*,q). 
\end{equation}

In video $v_*$, only the segment of the target moment $m_*$ holds relevance to the query. As a result, many prior methods typically adopt frame-aware retrieval for VR. We introduce a simple yet effective event-aware retrieval model for VCMR.

\subsection{Feature Extractor}
\label{sec:feature_extraction}
The initial features of model are extracted by pre-trained networks.  The visual features~(frame features) of video are encoded by 2D and 3D CNNs, i.e., ResNet~\cite{he2016deep} and Slowfast~\cite{feichtenhofer2019slowfast} to extract semantic and action features respectively.  The textual features of subtitles in video and text query are encoded by RoBERTa~\cite{liu2019roberta}. In particular, the feature of a frame in the video is obtained by max-pooling the visual features over a short duration~(1.5 seconds), and if subtitles are available at the corresponding time, it is featured as max-pooling word features in the subtitle at the corresponding duration. The visual features of frames in the $i$-th video $v_i$ is formulated as $\bm{F}_i = \{\bm{f}_i^1, \bm{f}_i^2, ..., \bm{f}_i^T\}$, and the subtitle features are  $\bm{S}_i = \{\bm{s}_i^1, \bm{s}_i^2, ..., \bm{s}_i^T\}$. If the subtitle is not available at a  time in the video, the corresponding text feature is a vector of zeros. The query feature is $\bm{Q} = \{\bm{w}^1, \bm{w}^2, ...,\bm{w}^L \}$. In this paper, we use bold symbols for vectors, distinguishing normal symbols such as $v_i$ that indicate a video. Before being fed into the model, all features are mapped by the fully connected layers to a space of the dimension $D$.

\subsection{Event-aware Video Retriever}
\label{sec:video_retriever}
We propose a two-tower event-aware retriever that utilizes the event representations of the videos as the retrieval units. The extraction of event representations involves event reasoning and hierarchical event encoding shown in~\Cref{fig:retriever}.

\subsubsection{Event Reasoning}
\label{sec:event_extraction}
We segment the video into units perceived by humans as events, emphasizing the gathering of consecutive and visually similar frames to form events. A representative work for event extraction is UBoCo~\cite{kang2022uboco} which leverages contrastive convolution to identify event boundaries.  We draw on this approach but simplify the process to make it more adaptable to VCMR. In addition, we also adopt two extra event extraction strategies, K-means and window.

\noindent \textbf{Contrastive convolution} Utilizing frame representations $\bar{\bm{F}} = \{\bar{\bm{f}}^1, \bar{\bm{f}}^2, ..., \bar{\bm{f}}^T \}$, we compute self-similarities among frames, thereby constructing a Temporal Self-Similarity Matrix~(TSM) shown in ~\Cref{fig:retriever}. A contrastive kernel is employed to perform convolution along the diagonal of TSM, for computing event boundary scores.
The results of diagonal elements serve as boundary scores, where a higher score indicates a greater likelihood that the frame is a boundary used to split video into events. We use a threshold $\delta$ to decide whether the i-$th$ frame is a boundary if the difference between the score and the mean of all scores is greater than $\delta$. 

\noindent \textbf{Kmeans} We employ TSM column vectors as features for K-means clustering, partitioning the video into $k$ segments to represent distinct events. To ensure consecutiveness within each segment, we include the frame index as an additional feature.

\noindent \textbf{Window} A fixed-size window divides the video evenly into pieces as events. The window size $w$ is a hyper-parameter.

This paper focuses on extracting visual events and still employing a frame-aware approach for subtitles. Extracting textual events poses more challenges as subtitle information is non-continuous, and subtitles with high similarity may not belong to the same event, as observed in the topic model~\cite{larochelle2012neural}. The extraction of textual events can be left for future works. 

\subsubsection{Hierarchical Event Encoding}

We employ a hierarchical structure to encode event representations, initially focusing on frame representation and subsequently on the event, ensuring the interactions of contextual information at both levels. Transformers for frame and event are augmented with anchor attention, encouraging them to focus on the correlations of neighboring content. And the video retriever is trained by two-branch contrastive learning.

\noindent \textbf{Anchor Attention} Untrimmed video contains abundant information, where not all frames or events exhibit strong correlations, and there is a tendency for higher correlations within close ranges. To this end, we introduce anchor multi-head self-attention~(AMHSA) to enhance the relevance between neighboring content.

We review  vanilla multi-head self-attention~(MHSA)~\cite{vaswani2017attention}: 
\begin{equation}
{\rm MultiHead}(Q,K,V) =  {\rm Concat}({\rm head}_1,...{\rm head}_h), 
\end{equation}
\begin{equation}
{\rm head}_i = {\rm Attention}(Q,K,V), 
\end{equation}
\begin{equation}
 \alpha = \frac{QK^t}{\sqrt{D}},\ \ {\rm Attention} = {\rm softmax}(\alpha) V,
\end{equation}
where $\alpha$ is the attention score before softmax normalized. In each attention head, an element in the input computes attention scores with all elements. Instead, we introduce a constraint, allowing an element to calculate attention scores only with a finite number of its neighboring elements. For instance, for the $i$-th frame, attention score computation is limited to the 2 frames before and after, forming a range of [i-2, i+2]. Different attention heads can utilize various ranges, such as 2, 3, 4, or all frames (ensuring globality) shown in~\Cref{fig:retriever}, capturing multi-scale neighborhood correlations. These ranges for attention computation serve as anchors, leading us to term it "anchor attention." We use AnchorFormer to mark the Transformer enhanced with anchor attention.

\begin{figure}
\begin{minipage}{0.5\textwidth}
\centering
\includegraphics[width=8.5cm]{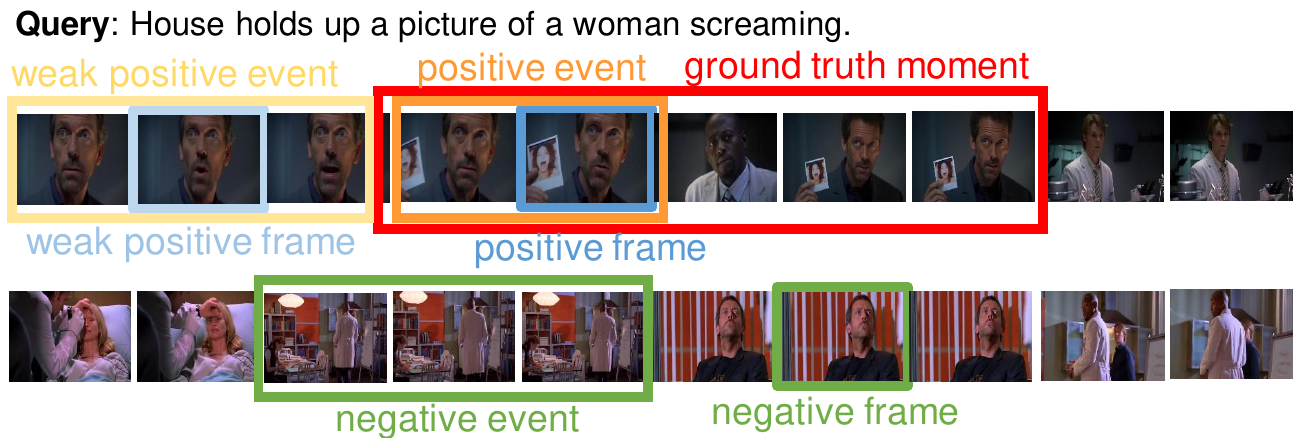}
\subcaption{Two-branch sampling for video retriever.}
\label{fig:contastivie_sampling}
\end{minipage}
\begin{minipage}{0.5\textwidth}  
\centering
\includegraphics[width=8.5cm]{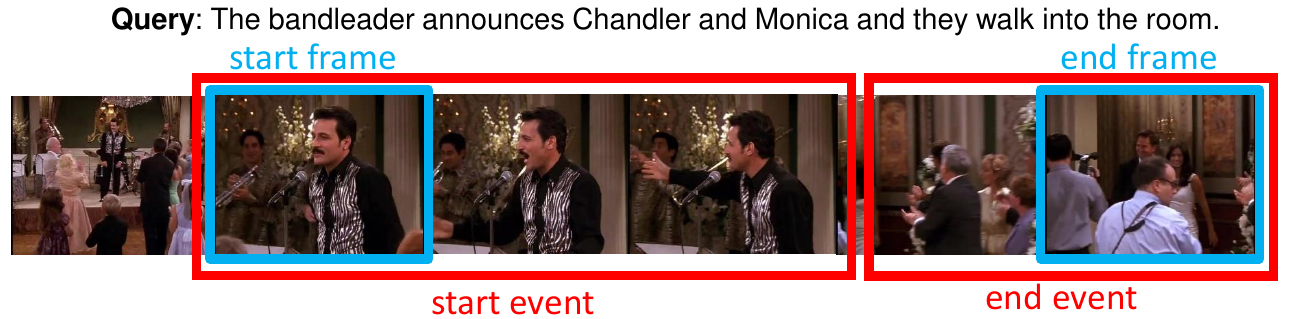}
\subcaption{Dual optimization for moment localizer.}
\label{fig:dual_optimization}
\end{minipage}
\centering
\caption{Two-branch sampling and dual optimization.}
\end{figure}

\noindent \textbf{Hierarchical Video Encoder}
The hierarchical encoding of the event involves frame and event encodings. We first encode frame representations.
For the $i$-th video, we input visual features of the video and textual features of subtitles, along with positional embeddings and modality embeddings, into a multi-modal AnchorFormer. This allows for the simultaneous capture of both intra-modal and inter-modal contextual dependencies. The output contextual representation of visual and textual modalities are $\bar{\bm{F}}_i = \{\bar{\bm{f}}_i^1, \bar{\bm{f}}_i^2, ..., \bar{\bm{f}}_i^T \}$ and $\bar{\bm{S}}_i = \{\bar{\bm{s}}_i^1, \bar{\bm{s}}_i^2, ..., \bar{\bm{s}}_i^T \}$ respectively. 

After event reasoning, we partition the video into $N$ events,
the initial event representation $\bar{\bm{E}}_i = \{\bar{\bm{e}}_i^1, \bar{\bm{e}}_i^2, ..., \bar{\bm{e}}_i^N \}$ is obtained from max pooling of the frame representations contained in event. Considering that events carry richer semantic information compared to frames, and the frequent presence of tight semantic associations between events, we employ an additional AnchorFormer to capture contextual dependencies at event level.
The input of AnchorFormer is event representations $\bar{\bm{E}}_i$ and subtitle representations $\bar{\bm{S}}_i$, and the output is contextual representations $\hat{\bm{E}}_i = \{\hat{\bm{e}}_i^1, \hat{\bm{e}}_i^2, ..., \hat{\bm{e}}_i^N \}$ and $\hat{\bm{S}}_i = \{\hat{\bm{s}}_i^1, \hat{\bm{s}}_i^2, ..., \hat{\bm{s}}_i^T \}$.

\noindent \textbf{Query Encoder}
Token features of a query are processed through vanilla Transformer to yield token representations $\bar{\bm{w}}^j$. Given the inconsistent matching of words across modalities in a query, as the query in~\Cref{fig:retriever}, where "...turns and leaves the room" emphasizes the visual modality with its action description, while "Foreman gives Chase a negative answer to his question..." leans towards the textual modality, we adopt modality-specific pooling to create two query representations for two modalities, denoted as $\bm{Q}_F$~(frame) and $\bm{Q}_S$~(subtitle).
Specifically, we calculate the weight of each word for a modality, followed by a weighted sum of word representations: 
\begin{equation} 
o^j = \bm{W}_d \bar{\bm{w}}^j,  \ \ \  \\
\ \alpha^j = \frac{{\rm exp}(o^j)}{\sum \limits_{i=1}^{L} {\rm exp}(o^i)}, \ \ \  \\
\ \bm{q}^d =  \sum \limits_{j=1}^{L} \alpha^j \bar{\bm{w}}^j  , 
\end{equation}
where $\bm{W}_d \in \mathbb{R}^{D \times 1}$ is a fully-connect layer which outputs a scalar $o^j$, $d \in \{F, S\}$ is frame or subtitle. $\alpha^j$ is softmax normalized weight of $j$-th word. And $\bm{Q}_d$ is a modality-specific query representation.

\begin{figure}
    \begin{center}
    \includegraphics[width=0.35\textwidth]{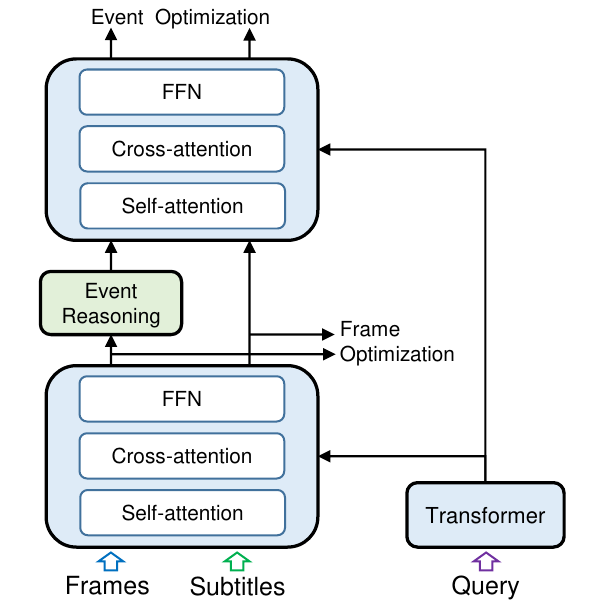}
    \end{center}
    \caption{Moment Localizer: both frame outputs and event outputs undergo optimization during training, while only frame outputs are utilized for prediction.}
    \label{fig:localizer}
\end{figure}

\begin{table*}[]
\caption{VR results on TVR validation set, ANetCaps validation set, and DiDeMo test set. $\dag$: the fine-tuned model before pre-training. The hyper-parameters~($\delta$, $k$, $w$) of the three event extraction strategies are set to (0.3, 10, 5), (0.1, 7, 8), and (0.1, 5, 4) for three datasets respectively. The results of XML and ReLoCLNet are reproduced by us using the same features.}
\begin{tabular}{lcccccccccccc}
\hline
\multirow{2}{*}{Model}    & \multicolumn{4}{c}{TVR}  & \multicolumn{4}{c}{ANetCaps} & \multicolumn{4}{c}{DiDeMo} \\
                          & R@1 & R@5 & R@10 & R@100 & R@1   & R@5   & R@10   & R@100  & R@1  & R@5  & R@10 & R@100 \\
\hline
XML~\cite{lei2020tvr} (ECCV'20) & 18.52 & 41.36 & 53.15 & 89.59 & 6.14 & 20.69 & 32.45 & 75.92 & 6.23 & 19.35 & 29.95 & 74.16  \\
ReLoCLNet~\cite{zhang2021video} (SIGIR'21) & 22.63 & 46.54 & 57.91 & 90.65 & 6.66 & 22.18 & 34.07 & 75.59 & 5.53 & 18.25 & 27.96 & 71.42  \\
HERO~\cite{li2020hero} (EMNLP'20) & 19.44 & 42.08 & 52.34 & 84.94 & 4.70 & 16.77 & 27.01 &  67.42 & 5.11 & 16.35 & 33.11 & 68.38 \\
${\rm HERO}^{\dag}$~\cite{li2020hero} (EMNLP'20) & 29.01 & 52.82 & 63.07 & 89.91 & 6.46 & 21.45 & 32.61  & 73.00  & \textbf{8.46} & 23.43 & 34.86 & 75.36      \\
SQuiDNet~\cite{yoon2022selective} (ECCV'22) & \textbf{31.61} & - & \textbf{65.32} & - & - & - & - & - & - & - & - & -  \\
\hline
EventFormer (Frame) & 25.56 & 50.14 & 61.33 & 91.79 & 7.50 & 24.20 & 37.10 & 77.97 & 7.63 & 24.06 & 35.06 & 77.63 \\
EventFormer (Convolution) & 28.44 & \textbf{52.92} & 64.11 & \textbf{92.92} & 7.97 & 25.51 & 37.97 & 77.62 & 8.19 & 23.77 & 35.33 & 77.67  \\
EventFormer (Kmeans) & 27.51 & 52.80 & 64.01 & 92.54 & \textbf{8.36} & \textbf{26.03} & \textbf{38.42} & \textbf{78.00} & 7.99 & 24.08 & 35.61 & 77.57 \\
EventFormer (Window) & 27.59 & 52.50 & 64.07 & 92.38 & 8.16 & 25.76 & 37.93 & 77.96 & 8.39 & \textbf{25.39} & \textbf{35.76} & \textbf{77.72} \\
\hline
\end{tabular}
\label{tab:vr_results}
\end{table*}

\noindent \textbf{Two-branch Contrastive Learning}
We introduce a two-branch contrastive learning method focusing on both frame and event representations for event representation learning. The additional frame representation learning aims to acquire more fitting representations for the query, considering that events are composed of frames. A key aspect in representation learning involves the selection of positive and negative samples~\cite{chen2020simple}. Shown in~\Cref{fig:contastivie_sampling}, we sample positive sample from the range of the correct moment, as this part is explicitly relevant to the query. Given the contextual coherence of the video, content beyond the range of the moment might possess implicit relevance to the query, such as content preceding and following the moment. We also take positive sample from content excluding the target moment in the video. And the negative samples are from videos irrelevant to query.

Specially, in frame branch, the positive sample and weak positive sample are frames exhibiting the highest query similarity within and outside the target moment, respectively.
For negative frames, we employ the hardest sample mining technique~\cite{faghri2017vse++}, wherein the frame within each negative video exhibiting the highest similarity to the query is chosen as the negative sample. The sampling for subtitle is the same
as the frame. We apply InfoNCE~\cite{oord2018representation} loss, and take the positive sample as an example: 
\begin{equation}
\mathcal{L}^f =  -log \frac{{\rm exp}(rf^+ / t)}{{\rm exp}(rf^+ / t) + \sum\limits_{z = 1}^{n} {\rm exp}(rf^-/t)}, 
\end{equation}
where $t$ is the temperature set to 0.01, $n$ is the number of negative videos, and $rf^+$ is the average of cosine similarities of query and positive frame/subtitle :
\begin{equation}
rf^+ = \frac{1}{2} ({\rm cos}(\bm{Q}_F, \bar{\bm{f}}^+ ) + {\rm cos}(\bm{Q}_S, \bar{\bm{s}}^+) ),
\end{equation}
where subtitle similarity $ {\rm cos}(\bm{Q}_S, \bar{\bm{s}}^+) $ is optional. The computation of weak positive frame loss $\mathcal{L}^f_w$ is identical to that of the positive frame. In addition to query-to-frame loss, following most works on cross-modal retrieval that employ bidirectional loss, we incorporate frame-to-query loss $\mathcal{L}^q$. The bidirectional loss for frame branch is: 
\begin{equation}
\mathcal{L}_F = \mathcal{L}^f +  \omega * \mathcal{L}^f_w + \mathcal{L}^q,
\end{equation}
where $\omega$ is a hyper-parameter set to 0.5.

For event branch, positive event is the event that contains positive frame to hold the consistency of contrastive learning. And the negative events sampled similarly to those in the frame branch. The overall loss for two-branch contrastive learning is:
\begin{equation}
\label{equ:loss_retriever}
\mathcal{L} = \lambda * \mathcal{L}_F + \mathcal{L}_E ,
\end{equation}
where $\mathcal{L}_E$ is InfoNCE loss of event representation learning between query representations $\bm{Q}_F$/$\bm{Q}_S$ and event and subtitle representations $\hat{\bm{e}}$/$\hat{\bm{s}}$, and $\lambda$ is a hyper-parameter set to 0.8.

\begin{table*}[]
\caption{SVMR and VCMR results on TVR validation and test set. The results of the test set are obtained by submitting predictions to the evaluation system.  $*$: reproduced results. $\dag$: the two-stage models that use HERO as the video retriever.}
\begin{tabular}{lccccccccc}
\hline
                          & \multicolumn{1}{l}{} & \multicolumn{1}{l}{SVMR (val)} & \multicolumn{1}{l}{} & \multicolumn{1}{l}{} & \multicolumn{1}{l}{VCMR (val)} & \multicolumn{1}{l}{} & \multicolumn{1}{l}{} & \multicolumn{1}{l}{VCMR (test)} & \multicolumn{1}{l}{} \\
\multicolumn{1}{l}{Model} & \multicolumn{1}{l}{} & IoU=0.7                       & \multicolumn{1}{l}{} & \multicolumn{1}{l}{} & IoU=0.7                       & \multicolumn{1}{l}{} & \multicolumn{1}{l}{} & IoU=0.7                        & \multicolumn{1}{l}{} \\
\multicolumn{1}{c}{}      & R@1                  & R@10                          & R@100                & R@1                  & R@10                          & R@100                & R@1                  & R@10                           & R@100                \\
\hline
HAMMER~\cite{zhang2020hierarchical} (Arxiv'20) & -                    & -                             & -                    & 5.13                 & 11.38                         & 16.71                & -                    & -                              & -                    \\
SQuiDNet~\cite{yoon2022selective} (ECCV'22) & 24.74                & -                             & -                    & 8.52                & -                             & -                    & 10.09                & 31.22                          & 46.05                \\
XML~\cite{lei2020tvr} (ECCV'20)  & $13.05^*$                & $38.80^*$                         & $63.13^*$                & $2.91^*$                 & $10.12^*$                         & $25.10^*$                & 3.32                 & 13.41                          & 30.52                \\
ReLoCLNet~\cite{zhang2021video} (SIGIR'21)  & $14.80^*$                & $45.85^*$                         & $72.39^*$                & $4.11^*$                 & $14.41^*$                         & $32.94^*$                & -                    & -                              & -                    \\
HERO~\cite{li2020hero} (EMNLP'20) & 15.30                & 40.84                         & 63.45                & 5.13                 & 16.26                         & 24.55                & 6.21                 & 19.34                          & 36.66                \\
${\rm CONQUER}^{\dag}$~\cite{hou2021conquer} (MM'21) & 22.84                & $53.98^*$                             & $79.24^*$                    & 7.76                 & 22.49                         & 35.17                & 9.24                 & 28.67                          & 41.98                \\
${\rm DMFAT}^{\dag}$~\cite{zhang2023video} (TCSVT'23)& 23.26                & -                             & -                    & 7.99                 & 23.81                         & 36.89                & -                    & -                              & -                    \\
${\rm CKCN}^{\dag}$~\cite{chen2023cross} (TMM'23) & 23.18                & -                             & -                    & 7.92                 & 22.00                         & 39.87                & -                    & -                              & -                    \\
\hline
EventFormer               & \textbf{25.45}                & \textbf{62.87}                         & \textbf{80.41}                & \textbf{10.12}                 & \textbf{27.54}                         & \textbf{42.88}                &              \textbf{11.11}  &          \textbf{32.78}     &     \textbf{46.18}            \\
\hline
\end{tabular}
\label{tab:vcmr_tvr}
\end{table*}

\subsection{Event-aware Moment Localizer}
\label{sec:moment_localizer}
The moment localizer shown in~\Cref{fig:localizer} is focused on accurately pinpointing the location of the target moment. We follow the works~\cite{lei2020tvr, zhang2021video, hou2021conquer, li2020hero} on VCMR of using proposal-free method, i.e., directly learning to predict the start and end positions of moment. We incorporate event information to proposal-free method, enhancing the model's learning to discriminate the start and end positions of moments through dual optimization of frame and event.

\noindent \textbf{Architecture} We introduce a one-tower event-aware moment localizer that has a similar structure to the retriever and also leverages AMHSA and event reasoning, but the video encoding requires cross-attention with the query. The query encoder is the same as in the video retriever, employing the vanilla Transformer to encode query words. However, the distinction is that the overall query representation $\bm{Q}_F$ or $\bm{Q}_S$ is unnecessary.  We emphasize that the architecture of the localizer is not novel; however, our innovation lies in the utilization of event and dual optimization.

\noindent \textbf{Dual Optimization} As shown in~\Cref{fig:dual_optimization}, for frame optimization, the objective is to maximize the confidence scores of frames which is the start or end boundary of ground truth moment. The confidence scores are derived from the output of AnchorFormer. Concretely, we begin by summing the visual and textual outputs at the same index, creating a sequence of multi-modal features with a length of $T$. Subsequently, the features are fed to two different 1D convolution networks to generate confidence scores for start $cf^{st} \in \mathbb{R}^1$ and end $cf^{ed}$ boundaries respectively. The 1D convolutions are used to capture dependencies among neighboring frames. The optimization is based on cross-entropy loss:
\begin{equation}
\label{equ:st_ed}
\mathcal{L}_F^{st} = -log   \frac{{\rm exp}(lf^{st})}{\sum \limits_{i=1} \limits^{T} lf_1^i}  , \ \ \ 
\mathcal{L}_F^{ed} = -log   \frac{{\rm exp}(lf^{ed})}{\sum \limits_{i=1} \limits^{T} lf_2^i}  , 
\end{equation}
\begin{equation}
\label{equ:l_f}
\mathcal{L}_F = \mathcal{L}_F^{st} + \mathcal{L}_F^{ed}, 
\end{equation}
where $lf_1^i$ and $lf_2^i$ are the $i$-th outputs from two convolution networks. For event optimization, we expect high confidence scores for events that contain correct moment boundaries. To obtain the confidence of the event, we use the output event representations and subtitle representations as features.  Similar to frame optimization, we also need text features for event prediction. We perform max-pooling on the subtitle representations within the scope of an event as the textual event features for the event. The sum of visual and textual features of an event are fed to two distinct fully connected networks to predict confidence scores that the moment boundaries are in the event. The optimization is the same as that in~\cref{equ:st_ed} and~\cref{equ:l_f}. The overall loss is:
\begin{equation}
\label{equ:loss_localizer}
\mathcal{L} = \mathcal{L}_F  + \gamma * \mathcal{L}_E,
\end{equation}
where $\mathcal{L}_E$ is event loss, $\gamma$ is a hyper-parameter set to 0.8.

\subsection{Training and Inference}
\label{sec:train_inference}

We employ a stage-wise training strategy for the two modules. Firstly, the video retriever is trained using an in-batch negative sampling method~\cite{karpukhin2020dense}, where all other videos in a batch serve as negative videos. Subsequently, the moment localizer is trained using sharing normalization techniques~(Shared-Norm)\cite{clark-gardner-2018-simple}, widely applied in open domain QA\cite{chen-etal-2017-reading} tasks. This technique enhances the confidence that the moment appears in the correct video while reducing its confidence in the wrong video. Especially, the softmax normalizations in the loss functions \cref{equ:st_ed} cover confidence scores not only for frames or events in the correct video but also in incorrect videos, serving as negative samples. The negative videos are sampled from the training set based on high similarity to the query, with the similarity computed by the trained video retriever.

In inference, we first use the video retriever to retrieve the top-10 videos from the corpus based on the the average of the highest query-event and query-subtitle similarities $re_i$ in the video $v_i$. The moment localizer is used to predict the position of the moment in the 10 videos, relying on the confidence scores ($lf^{st}_i$ and $lf^{ed}_i$) indicating whether a frame serves as a start or end boundary.  The event aspect of moment localizer is excluded from the prediction, as moment localization necessitates fine-grained frame-level localization. The confidence score $cm$ for moment prediction consists of video retrieval score and moment localization score:
\begin{equation}
cm = \frac{re_i}{t} + lf^{st}_i + lf^{ed}_i,
\end{equation}
where $t$ is temperature in contrastive learning, consistent with the training objective, and $cm$ is used to rank the candidate moments.

\begin{table}[]
\caption{SVMR and VCMR results on ANetCaps validation set and DeDiMo test set. The metric is $R@1, IoU = 0.5,0.7$. }
\begin{tabular}{llcccc}
\hline
\multirow{2}{*}{Dataset}  & \multirow{2}{*}{Model}           & \multicolumn{2}{c}{SVMR}                                & \multicolumn{2}{c}{VCMR}                              \\
                          &                                  & 0.5                        & 0.7                        & 0.5                       & 0.7                       \\
\hline
\multirow{5}{*}{ANetCaps} & HAMMER~\cite{zhang2020hierarchical}    & 41.45                      & 24.27                      & 2.94                      & 1.74                      \\
                          & ReLoCLNet~\cite{zhang2021video}  & -                          & -                          & 3.09                      & 1.82                                            \\
                          & CONQUER~\cite{hou2021conquer}    &    $35.63^*$                &     $20.08^*$          &   $2.14^*$      &     $1.33^*$             \\ \cline{2-6} 
                          & EventFormer & \textbf{45.21} & \textbf{27.98} & \textbf{4.32} & \textbf{2.75} \\ 
\hline
\multirow{6}{*}{DiDeMo}   & XML~\cite{lei2020tvr} & -                          & -                          & 2.36                      & 1.59                      \\
                          & ReLoCLNet~\cite{zhang2021video} & $34.81^*$    &  $26.71^*$     & $2.28^*$                      & $1.71^*$  \\
                          & HERO~\cite{li2020hero}   &   $\textbf{39.20}^*$  &  $30.19^*$  & $3.42^*$                      & $2.79^*$                      \\
                          & CONQUER~\cite{hou2021conquer} & 38.17                      & 29.9                       & 3.31                      & 2.79                      \\
                          & DMFAT~\cite{zhang2023video}  & -                          & -                          & 3.44                      & 2.89                      \\
                          & CKCN~\cite{chen2023cross} & 36.54                      & 28.89                      & 3.22                      & 2.69                      \\ \cline{2-6} 
                          & EventFormer & 39.02 & \textbf{30.91} & \textbf{3.53} & \textbf{3.12} \\ 
\hline
\end{tabular}
\label{tab:vr_act_didemo}
\end{table}

\section{Experiments}
\subsection{Experimental Details}
\noindent \textbf{Datasets} We evaluate EventFormer on three benchmarks.  TV shows retrieval~(\textbf{TVR})~\cite{lei2020tvr} is constructed on TV shows with videos providing subtitles. The training, validation, and testing sets of TVR consist of 17,435, 2,179, and 1,089 videos, respectively. Each video contains 5 moments for retrieval. The average duration of the videos and moments are 76.2 seconds and  9.1 seconds respectively. ActivityNet Captions~(\textbf{ANetCaps})~\cite{caba2015activitynet} comprises approximately 20K videos. The videos exclusively contain visual information without subtitles. We follow the setup in~\cite{zhang2020hierarchical, yoon2022selective} with 10,009 videos for training and 4,917 videos for testing, resulting in 37,421 and 17,505 moments respectively. The average duration of videos and moments are 120 seconds and 36.18 seconds respectively. The videos of Distinct Describable Moments~(\textbf{DiDeMo})~\cite{anne2017localizing} are from YFCC100M~\cite{thomee2016yfcc100m}, exclusively feature visual information. The dataset is divided into 8,395, 1,065, and 1,004 videos for training, validation, and testing, respectively. Most videos have a duration of approximately 30 seconds, uniformly segmented into 5-second intervals, resulting that moment boundaries consistently aligning with multiples of 5.

\noindent \textbf{Implementation} For TVR and DiDeMo, we utilize the 768D RoBERTa feature provided by~\cite{lei2020tvr} for query and subtitle, and the 4352D SlowFast+ResNet feature provided by~\cite{li2020hero} as the frame feature. The duration for the sampling frame feature is 1.5 seconds with an FPS of 3. We follow the feature extractions in~\cite{lei2020tvr} and~\cite{li2020hero} to extract features for ANetCaps. In inference, we first retrieve the top-10 videos, then localize the moment within the retrieved videos. Non-maximum suppression~\cite{girshick2014rich} is applied in moment localization to remove overlapped predictions. For Shared-Norm at the moment localizer, the number of negative videos is set to 5, sampled from the top-100 videos in the training set, ranked by the video retriever. Anchor sizes for AnchorFormer in frame and event are configured as {3, 6, 9, all} and {1, 2, 3, all}, respectively.

\noindent \textbf{Evaluation Metrics} Following~\cite{lei2020tvr}, the metrics for VR are the same as those for text-to-video retrieval, i.e., $R@K$~($k=1,5,10,100$) the fraction of queries that correctly retrieve correct videos in the top K of the ranking list. And for SVMR and VCMR, the metrics are $R@K, IoU=\mu$~($\mu=0.5,0.7$) which require the intersection over union~(IoU) of predicted moments to ground truth exceeds $\mu$. The evaluation of SVMR is only in the correct video for a query, while the evaluation of VCMR ranges over videos in the corpus.

\noindent \textbf{Baselines} We compare our model to the  models for VCMR task as a baseline, containing one-tower methods, two-tower methods, and two-stage methods. One-tower: HAMMER~\cite{zhang2020hierarchical}, SQuiDNet~\cite{yoon2022selective}.   Two-tower: XML~\cite{lei2020tvr}, ReLoCLNet~\cite{zhang2021video}, HERO~\cite{li2020hero}. Two-stage: CONQUER~\cite{hou2021conquer}, DMFAT~\cite{zhang2023video}, CKCN~\cite{chen2023cross}. Additionally, a frame-aware baseline, denoted as EventFormer~(Frame), is introduced for VR. This baseline shares the same model architecture and the number of parameters as the proposed EventFormer but lacks anchor attention, event reasoning and event encoding modules.

\subsection{Main Results}

\noindent \textbf{VR}
The results of VR task on three datasets are reported in~\Cref{tab:vr_results}.  Except for the one-tower model SQuiDNet and the pre-trained large video-language model HERO, our event-aware retrieval model surpasses other frame-aware retrieval models such as XML, ReLoCLNet, and the frame-aware version of EventFormer. SQuiDNet leverages fine-grained cross-modal interaction between video and query for better matching. Nevertheless, the one-tower method encounters retrieval efficiency challenges as it involves fine-grained interactive matching between the query and each video in the corpus. HERO is pre-trained on HowTo100M~\cite{miech2019howto100m} and TVR, providing external knowledge for retrieval. However, HERO's performance on ANetCaps and DiDeMo is sub-optimal, likely due to a domain gap between the videos in the two datasets and HERO's pre-training data. For three event strategies, convolution surpasses the other two strategies in TVR, demonstrating superior adaptability to varying numbers of events in videos and dynamic event spans. Kmeans performs better in ANetCaps, attributed to the fact that the majority of videos in ANetCaps come from YouTube, which are user-shot, one-shot, continuous sequences, distinct from TV shows with explicit scene transitions. In DiDeMo, the window strategy excels because of the consistently fixed size of the query-related portion in videos, as detailed in~\cite{anne2017localizing}. Future works can be the exploration of more robust extraction methods for diverse datasets.

\begin{table}[]
\caption{Ablation of video retriever on TVR validation set. 'S': subtitle. 'ER': event reasoning. 'EI': event interaction. 'AMHSA': anchor multi-head self-attention. 'FCL': frame contrastive learning. 'ECL':event contrastive learning. 'WP': weak positive sample.}
\scalebox{0.81}{
\begin{tabular}{cccccccllll}
\hline
S & ER & EI & AMHSA & FCL & ECL & WP & R@1   & R@5   & R@10  & R@100 \\
\hline
 $\surd$ & $\surd$ & $\surd$ & $\surd$ & $\surd$ & $\surd$ & $\surd$ & 28.44 & 52.92 & 64.11 & 92.92 \\
         & $\surd$ & $\surd$ & $\surd$ & $\surd$ & $\surd$ & $\surd$ & 17.15 & 38.84 & 50.30 & 86.97 \\
 $\surd$ &         &         & $\surd$ & $\surd$ &         & $\surd$ & 25.56 & 50.14 & 61.33 & 91.79 \\
 $\surd$ & $\surd$ &         & $\surd$ & $\surd$ & $\surd$ & $\surd$ & 26.54 & 51.43 & 62.01 & 91.91 \\
 $\surd$ & $\surd$ & $\surd$ &         & $\surd$ & $\surd$ & $\surd$ & 26.29 & 51.55 & 62.18 & 92.13 \\
 $\surd$ & $\surd$ & $\surd$ & $\surd$ & $\surd$ &         & $\surd$ & 26.84 & 51.48 & 62.55 & 91.90 \\
 $\surd$ & $\surd$ & $\surd$ & $\surd$ & $\surd$ & $\surd$ &         & 27.49 & 52.53 & 64.01 & 92.69 \\
\hline  
\end{tabular}}
\label{tab:ablation_vr}
\end{table}

\begin{table}[]
\caption{Ablation of moment localizer on TVR validation set. 'EO': event optimization.'SN': Shared-Norm.}
\scalebox{0.82}{
\begin{tabular}{ccccllll}
\hline
\multicolumn{1}{c}{\multirow{2}{*}{S}} & \multicolumn{1}{c}{\multirow{2}{*}{AMHSA}} & \multicolumn{1}{c}{\multirow{2}{*}{EO}} & \multicolumn{1}{c}{\multirow{2}{*}{SN}} & \multicolumn{2}{c}{SVMR (R@1)}                          & \multicolumn{2}{c}{VCMR(R@1)}                          \\
\multicolumn{1}{c}{}                     & \multicolumn{1}{c}{}                       & \multicolumn{1}{c}{}                       & \multicolumn{1}{c}{}                    & \multicolumn{1}{c}{0.5} & \multicolumn{1}{c}{0.7} & \multicolumn{1}{c}{0.5} & \multicolumn{1}{c}{0.7} \\
\hline
$\surd$ & $\surd$ & $\surd$ & $\surd$ & 47.14       & 25.45      & 17.79       & 10.12       \\
        & $\surd$ & $\surd$ & $\surd$ & 41.75       & 22.07      & 14.80       & 8.00       \\
$\surd$ &         & $\surd$ & $\surd$ & 44.64       & 23.89      & 16.49       & 9.23       \\
$\surd$ & $\surd$ &         & $\surd$ & 46.25       & 25.12      & 17.12       & 9.80       \\
$\surd$ & $\surd$ & $\surd$ &         & 43.52       & 23.07      & 14.14       & 8.14      \\
\hline               
\end{tabular}}
\label{tab:ablation_svmr}
\end{table}

\noindent \textbf{SVMR and VCMR}
The results of SVMR and VCMR on three datasets are reported in~\Cref{tab:vcmr_tvr} and~\Cref{tab:vr_act_didemo}. In both tasks, our proposed EventFormer outperforms other baselines, no matter which architectures~(one-tower, two-tower, and two-stage) these models belong to. The one-tower~(HAMMER, SQuiDNet) and two-stage~(CONQUER, DMFAT, and CKCN) models exhibit superior performance compared to the two-tower~(XML, ReLoCLNet, HERO) models. This is attributed to fine-grained interactions, making deep matching between the query and video. 
In contrast, the two-tower model relies solely on the similarity between frames and queries to determine the boundaries of the target segments.
EventFormer has a similar structure to the other two-stage models, leveraging Transformer for multi-modal fusion. However, our model surpasses these models, attributed to AMHSA and dual optimization.

\subsection{Ablation Study}
The results on TVR validation set for video retriever and moment localizer are reported in~\Cref{tab:ablation_vr} and~\Cref{tab:ablation_svmr}, respectively. 

\noindent \textbf{Video retriever} Subtitle plays a crucial role, as many queries in TVR include character names like "Shelton", which align more effectively with the textual information than with visual content. The retrieval accuracy significantly benefits from event reasoning, event interaction, and anchor attention, thus validating the three reasons highlighted in the Introduction for the ineffectiveness of frame-aware methods in leveraging event information for video retrieval. Frame learning in two-branch contrastive learning works, demonstrating the query-related frame representations contribute to the learning of event representations. Moreover, weak positive sample enhance learning by taking implicitly query-related frame or event.

\noindent \textbf{Moment localizer} Subtitle information also helps, but the improvement is not as pronounced as in video retriever. This is because moment localization involves precisely identifying the action described by the query, placing more emphasis on visual information, with text typically playing a supporting role.  AMHSA is also effective for moment localization. Event optimization enhances retrieval accuracy, even without direct involvement in the prediction, indicating the beneficial impact of additional optimization. Notably, Shared-Norm exerts a substantial influence on moment localization, particularly in VCMR, as this technique empowers the model with the capability to distinguish moments in different videos.

\begin{table}[]
\caption{The results of moment localization directly using event extracted by three extract strategies.}
\scalebox{0.85}{
\begin{tabular}{lcccc}
\hline
\multirow{2}{*}{Strategy} & \multicolumn{2}{c}{SVMR} & \multicolumn{2}{c}{VCMR} \\
                          & 0.5         & 0.7        & 0.5         & 0.7        \\
\hline
Window ($w$ = 5)                   & 20.59       & 7.86       & 6.63        & 2.74       \\
Kmeans ($k$ = 10)                   & 21.15       & 9.03       & \textbf{7.11}        & 3.23       \\
Convolution ($\delta$ = 0.3)              & \textbf{21.31}       & \textbf{9.88}      & 6.91        & \textbf{3.51}  \\
\hline
\end{tabular}}
\label{tab:event_moment}
\end{table}

\subsection{Event Reasoning}

We evaluate three event extraction strategies on TVR validation set. Specially, we predict the moment by directly using the event with the highest similarity computed by video retriever to the query in the video. The results are presented in~\Cref{tab:event_moment}. While the accuracy falls short of the optimal events in the ideal case that is shown in~\Cref{tab:event_validation}, it still demonstrates effectiveness. This is because the events are extracted solely through the aggregation of consecutive and similar frames, without training for the localization task. The events extracted through contrastive convolution are closer to the ground truth moment compared to Kmeans and window for its superior adaptation to the number and length of events.

\begin{table}[]
\caption{Efficiency and memory on TVR validation set. Efficiency is measured as the average latency~(ms) to retrieve the top-10 videos. And memory~(MB) is the storage of vectors of frames or events saved in advance by two-tower model. 'Number' is the total number of frames or events in corpus.}
\scalebox{0.83}{
\begin{tabular}{lccccc}
\hline
\multirow{2}{*}{Model}       & \multicolumn{1}{l}{\multirow{2}{*}{Params}} & \multicolumn{3}{c}{VR}                                                                & \multicolumn{1}{l}{SVMR}    \\
            & \multicolumn{1}{l}{}                             & \multicolumn{1}{l}{Latency} & \multicolumn{1}{l}{Memory} & \multicolumn{1}{l}{Number} & \multicolumn{1}{l}{Latency} \\
\hline
CONQUER~\cite{hou2021conquer}    & 47M                                              & 9932                       & -                          & -                          & 156                         \\
ReLoCLNet~\cite{zhang2021video}         & \textbf{8M}                                               & 88                          & 161                        & 109924                     & \textbf{29}                          \\
HERO~\cite{li2020hero}  &  121M             & 212                         & 322                        & 109924                     & 97        \\
EventFormer ($\delta = 0.3$) & 9M+18M                                           & 51                          & 47                         & 31975                      & 103   \\
EventFormer ($k = 10$) & 9M+18M                                           & \textbf{43}                          & \textbf{31}                         & \textbf{21786}                      & 103   \\
EventFormer ($w = 5$) & 9M+18M        &    \textbf{43}           &  33                          &     22755      & 103    \\
\hline
\end{tabular}
}
\label{tab:efficiency_memory}
\end{table}

\subsection{Retrieval Efficiency and Memory Usage}
We further analyze the retrieval efficiency and memory usage of our EventFormer and other models. We select a two-tower model ReLoCLNet, a two-tower pre-trained model HERO and a one-tower model. Given that the one-tower models HAMMER and SQuiDNet lack published code, we choose CONQUER due to its attempt at VR task introduced in~\cite{hou2021conquer}. The results are reported in~\Cref{tab:efficiency_memory}. In VR, CONQUER shows the slowest performance because it cannot decompose similarity, requiring online calculations for the relevance between query and videos. HERO exhibits lower efficiency compared to XML and our model, attributed to its excessive number of parameters and the representations with twice the dimensionality of XML and our model. Our model is optimally efficient and least memory consuming because the number of saved events is much smaller than the number of frames. In SVMR, although our model is not as efficient as two-tower models, it remains acceptable since only 10 videos interact with the query at a fine-grained level.

\begin{table}[]
\caption{PRVR results on TVR~(without subtitle) validation set. All models use the same features, ResNet+I3D~\cite{carreira2017quo} for video and RoBERTa for the query.}
\scalebox{0.85}{
\begin{tabular}{llllll}
\hline
Model         & R@1    & R@5    & R@10   & R@100   & SumR   \\
\hline
\multicolumn{6}{l}{\textit{VCMR models w/o moment localization: }} \\
\hline
XML~\cite{lei2020tvr} (ECCV'20)          & 10.0   & 26.5   & 37.3   & 81.3    & 155.1  \\
ReLoCLNet~\cite{zhang2021video} (SIGIR'21) & 10.7   & 28.1   & 38.1   & 80.3    & 157.1  \\
CONQUER~\cite{hou2021conquer} (MM'21) & 11.0   & 28.9   & 39.6   & 81.3    & 160.8  \\
\hline
\multicolumn{6}{l}{\textit{PRVR  models: }}                       \\
\hline
MS-SL~\cite{dong2022partially} (MM'22)  & 13.5   & 32.1   & 43.4   & 83.4    & 172.4  \\
PEAN~\cite{jiang2023progressive} (ICME'23) & 13.5   & 32.8   & 44.1   & 83.9    & 174.2  \\
GMMFormer~\cite{wang2023gmmformer} (AAAI'24)& 13.9   & 33.3   & 44.5   & 84.9   & 176.6  \\
DL-DKD ~\cite{dong2023dual} (ICCV'23)   & \textbf{14.4}   & \textbf{34.9}   & 45.8   & \textbf{84.9}    & \textbf{179.9}  \\
\hline
EventFormer~($\delta = 0.3$)   & 14.2   & 34.6   & \textbf{46.0}   & 84.8    & 179.6  \\
\hline
\end{tabular}
}
\label{tab:prvr}
\end{table}

\begin{figure}
    \begin{center}
    \includegraphics[width=0.5\textwidth]{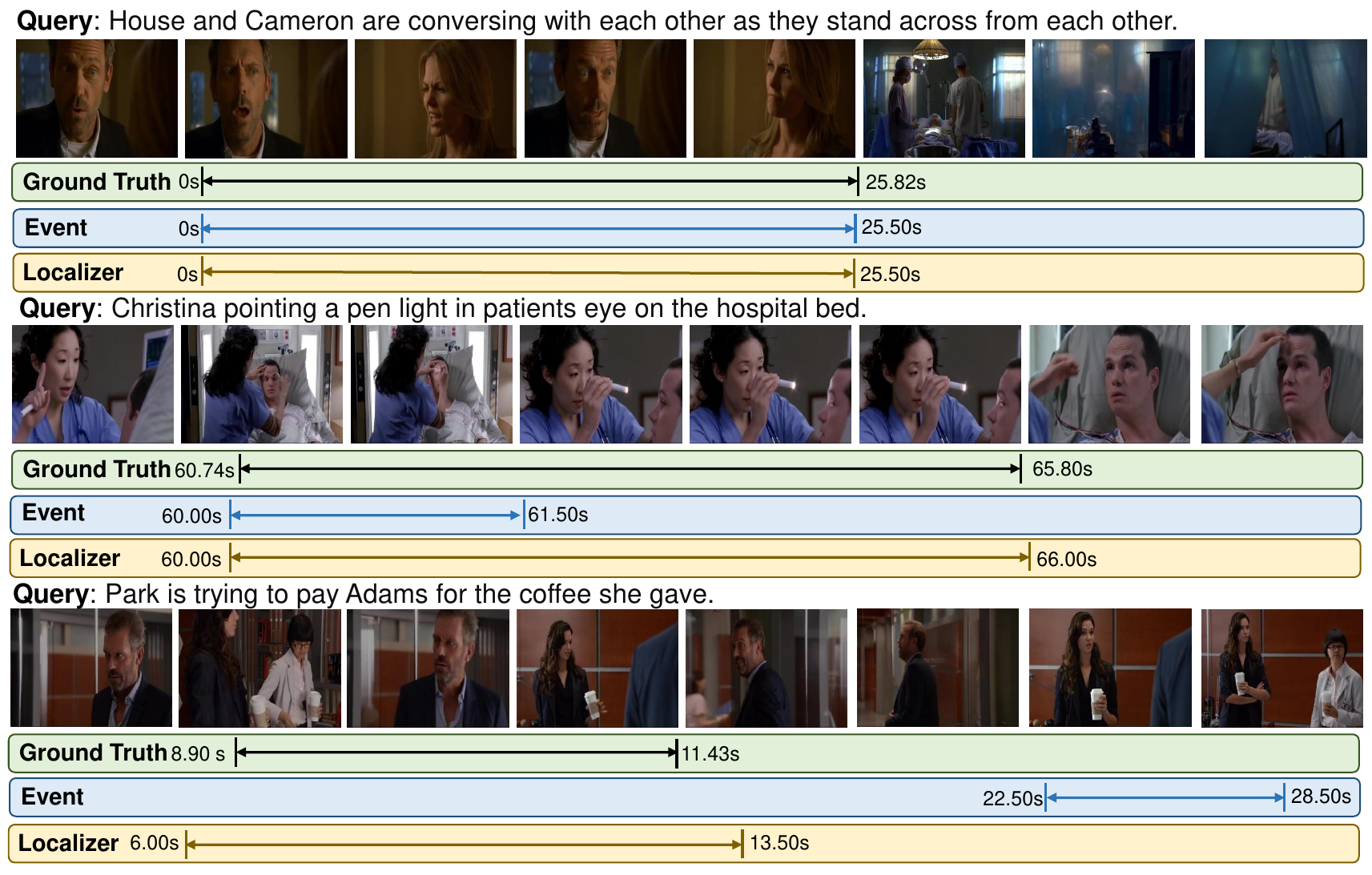}
    \end{center}
    \caption{Cases of events extracted by retriever and moments predicted by localizer.  }
    \label{fig:case}
\end{figure}

\subsection{Partially Relevant Video Retrieval}
Beyond VCMR, we evaluate the proposed EventFormer on the PRVR task, which serves as a weakly supervised version of VR, as it does not provide the ground truth moment for the query. This poses a challenge for our model because the positive frames and events for two-branch contrastive learning are sampled based on the moment relevant to the query. We modify the sampling strategy to adapt to PRVR, selecting the two frames or events in the correct video with the highest similarity to the query as positive and weak positive samples. The negative sampling from other videos are same as that of supervised EventFormer. The results are presented in~\Cref{tab:prvr}. Our model demonstrates superior retrieval accuracy compared to other models, except for DL-DKD, which distills knowledge from a large-scale vision-language model. Notably, our model achieves this performance while being trained solely on the dataset's training set. Our model is designed for supervised VCMR task, leaving room for enhancement in weakly supervised VR for feature works.

\subsection{Case Study}
We present three cases in~\Cref{fig:case}. In the first case, the extracted event overlaps perfectly with the ground truth, because the boundaries of the moment fall exactly where the visual content suddenly changes. In the second case, the change occurs in the middle of the moment, resulting in the event capturing the front half of the content. However, this part remains pertinent to the query. The last case is a failure example, where multiple changes occur within the moment, making the event reasoning struggle to capture consecutive and similar frames.

\section{Conclusion}
This paper proposes an event-aware retrieval model EventFormer for the VCMR task, motivated by human perception of visual information. To extract event representations of video for retrieval, EventFormer leverages event reasoning and two-level hierarchical event encoding. Anchor multi-head self-attention is introduced for Transformer to enhance close dependencies in the untrimmed video. We adopt two-branch contrastive learning and dual optimization for the training of two sub-tasks in VCMR. Extensive experiments show the effectiveness and efficiency of EventFormer on VCMR. The ablation study and case study additionally further verify the efficacy and rationale of each module in our model. The effectiveness of the model is also validated on the PRVR task. Our approach has limitations, particularly in robustness for videos in different datasets. Additionally, our event reasoning relies mainly on visual frame similarity, making it sensitive to changes in visual appearance. Future work can address these problems by introducing more semantic associations.

\bibliographystyle{ACM-Reference-Format}
\bibliography{sample-base}

\end{document}